# Small Boxes Big Data: A Deep Learning Approach to Optimize Variable Sized Bin Packing

Feng Mao * Edgar Blanco *† Mingang Fu * Rohit Jain * Anurag Gupta* Sebastien Mancel *
Rong Yuan * Stephen Guo * Sai Kumar * Yayang Tian*

* *Warlmart Labs & Walmart Global eCommerce*
*Sunnyvale, CA, 94086 & San Bruno, CA 94066 USA*
{*fmao,eblanco,mfu,rjain,agupta10,smancel,ryuan,sguo,skumar2,yayang*}@walmartlabs.com
*fengmao@acm.org*

† *MIT Center for Transportation & Logistics*
*Cambridge, MA, 02142, USA*
*eblanco@mit.edu*

*Abstract*—Bin Packing problems have been widely studied because of their broad applications in different domains. Known as a set of NP-hard problems, they have different variations and many heuristics have been proposed for obtaining approximate solutions. Specifically, for the 1D variable sized bin packing problem, the two key sets of optimization heuristics are the bin assignment and the bin allocation. Usually the performance of a single static optimization heuristic can not beat that of a dynamic one which is tailored for each bin packing instance. Building such an adaptive system requires modeling the relationship between bin features and packing perform profiles. The primary drawbacks of traditional AI machine learnings for this task are the natural limitations of feature engineering, such as the curse of dimensionality and feature selection quality. We introduce a deep learning approach to overcome the drawbacks by applying a large training data set, auto feature selection and fast, accurate labeling. We show in this paper how to build such a system by both theoretical formulation and engineering practices. Our prediction system achieves up to 89% training accuracy and 72% validation accuracy to select the best heuristic that can generate a better quality bin packing solution.

*Keywords*-1D bin packing, deep learning, big data and optimization

## I. INTRODUCTION

Bin packing problems have been widely studied [1] among many application domains including logistics transportation, job scheduling, manufacturing, financial stock decisions, computer memory management [2], cloud resource assignment [3], network optimization like WiMax, etc. As a set of NP hard problems [4], they have many variations dependent upon the constraints for the specific problem. A typical 1D bin packing problem [2] packs a set of segments into another set of segments. A 3D bin packing finds a solution to pack a set of boxes into a set of boxes. According to the number of different candidate bin types, bin packing problems are divided into single sized bin packing and variable sized bin packing [5], which is commonly seen in logistics delivery packing.

Many approximation algorithms [1] have been proposed to find sub-optimal solutions in a short response time. In most cases, it is a trade off between the solution quality and the time to solve it. Adapting to different problem instances by different algorithms has been studied by traditional AI machine learning approaches [6] [7] [8] [9]. These works initially optimize the packing algorithm end-to-end and later to a finer granularity, state-to-state. To apply traditional machine learning, these works apply certain ad-hoc feature engineering to derive bin packing features for their learning algorithms.

The limitation of these approaches is that they heavily rely on feature engineering. In most cases, better quality data is more important than the model itself. To traditional machine learning, reliance on the feature engineering has two consequences: limitations from the curse of dimensionality [10] and the incompleteness and inaccuracy of extracted features. Deep learning [11] can mitigate these issues, although it is initially limited by the amount of available training data and the computation power required. In the current era of big data and multi-core computing, the amount of data we collect is large and the commodity computer power is sufficient enough for supporting the deep learning. This work proposes a deep learning approach for solving the bin packing problem. In this paper, we limit our scope to the 1D variable sized bin packing.

The challenges of the deep learning approach to the 1D variable sized bin packing lie in three folds. The first challenge comes from the problem definition itself. We need to explore and represent the optimization heuristic space for the algorithm formally. Second, from the data perspective, it is hard to label a large set of instance performance profiles which overlap with each other within a practical time constraint. Third, in terms of model building, it is difficult to determine the optimal neural network structure and its parameters, such as depth applied and activation function to use. All these factors need to cooperate to cancel noises but keep signals from a large set of raw features.

In this study, we formally define 1D variable sized bin packing problem and the heuristic space for its optimization. The key parts of the approximation algorithm are how to assign the opened bins to the incoming item and how to choose what type of bins to open when it is required. We formally construct a set of heuristics for the assignment and allocation operation. We treat the optimization problem as a supervised learning procedure. To build a classification model, we need to label the performance profile. Instead of using the statistical significance of the performance profile classification [12], we propose the performance signature, a binary vector and a fast prefix based clustering algorithm. We further optimize the clustering by minimizing clustering entropy via Monte Carlo simulation [13]. We explored different neural network structures and parameters to tune the prediction model. We explored different depths of 2 to 16 levels and widths of 16 nodes to 512 nodes. We also explored 8 types of different activation functions from the traditional sigmoid to rectifier. Our experiments show that our system can achieve up to 89% training accuracy and 72% validating accuracy when choosing the best heuristic that generates a better quality bin packing solution.

The contributions of our work are listed as follows. First, we formalize and define the optimization space for 1D variable sized bin packing. Second, we propose a labeling procedure which balances the clustering time complexity and quality. Third, we mitigate feature engineering issues by exploring a deep learning approach and constructing the bin packing optimization prediction model.

The layout of this study is as follows. Section 2 describes the scope of the work through a brief survey of bin packing problems and a formal definition of 1D variable sized bin packing. Section 3 defines the heuristic space for 1D variable sized bin packing and the procedure of optimization. Section 4 presents the prediction model we use to predict the optimization decision, including the aspects of data collection, feature definition and instance labeling. Section 5 describes the predictive system as a high level, then illustrates how each component interacts with each other. Section 6 presents the structure of the neural network for deep learning and the engineering aspects such as over-fitting and speed-up. Section 7 contains the experiment results. Section 8 discusses related work. Section 9 concludes the paper with a short summary.

## II. BIN PACKING

The definition of the canonical bin packing problem is as follows: pack a finite set of items into a finite number of bins such that dimension constraints hold, while optimizing a cost function. Examples of common cost functions optimized include: minimization of number of bins and minimization of wasted bin space. In computational complexity theory, it is a combinatorial NP-hard problem [4]. As the search space is exponentially large, most practical solutions consider trade-offs between computational complexity and solution quality by finding a sub-optimal solution in a relatively short time using approximate heuristics.

### A. Bin Packing Classification

There are many variations of this problem and all of them are NP-hard. These problems have a wide range of applications such as optimization of logistics networks, semiconductor chip design, storage software optimization, manufacturer cost and financial stock decisions. They can be classified by different criteria [1].

The number of dimensions for the bins and items can be 1D, 2D or 3D. The memory management problem is a typical 1D bin packing problem [2]. Unlike the multi-dimension bin packing problem whose dimensions have dependency on the others, the vector bin pack packing has a vector of independent dimensions [3]. The cluster resource management is a typical application in this domain. Depending on whether the algorithm can see all the items beforehand, there are on-line [14] and off-line [15] bin packing problems. The number of different candidate bin types divides the bin packing problems into the single sized bin packing and the variable sized bin packing [5]. They are commonly seen in the pick-up cabinets and logistics delivery boxing. We summarize these variations in Table I.

Table I
BIN PACKING CLASSIFICATION

| Classification Criteria | Set of Types |
|---|---|
| Item/Bin Dimension | 1D, 2D and 3D |
| Vector | 1D and multiple D |
| Items Visibility | on-line/off-line |
| Number of Bin Types | single sized, variable sized |

In this work, we focus on the off-line one dimensional variable sized bin packing problem.

### B. 1D Variable Sized Bin Packing

Formally, given two sets of positive real numbers, $T = \{T_i \mid 1 \leq i \leq m\}$ and $B = \{B_i \mid 1 \leq i \leq n\}$ representing an item set and a bin set, we want to find a partition of $T$, set $TP = \{TP_i \mid 1 \leq i \leq k \leq m\}$ and a mapping $M \subset TP \times B$, $M = \{<TP_i, B_j> \mid (TP_i \in TP) \wedge (B_j \in B) \wedge (\sum_{l=1}^{|TP_i|} T_l \leq B_j, T_l \in TP_i)\}$, such that under the partition and mapping, the cost function $C = Cost(TP, M)$ can be minimized.

The cost function can be as simple as the size of a mapping. In this case, the cost function represents the total number of bins used. A more complex cost function can associate different costs with different bin types. One commonly used cost function is to define a unit space cost for each bin type and evaluate the waste of the bin space for a particular packing result. In this paper, we give each bin a fixed cost 1 and calculate each unit space cost by dividing its volume. The cost of a packing is defined as the wasted space cost $C = \sum_{m=1}^{|M|} 1 - volume(TP_i)/volume(B_j)$.

### III. HEURISTICS AND OPTIMIZATION DECISION

Two important decisions for the variable sized bin packing are bin assignment and allocation. Bin assignment determines how the item gets assigned to the existing allocated bins. Bin allocation determines what type of bin will get allocated when the item can not find a fit in the existing bins. We in this paper define a set of heuristics to make those decisions. Different assignment and allocation heuristics result in different costs, as defined by the cost function chosen. The set of heuristics form the search space where the optimization decision is made during a bin packing instance execution. The bin allocation heuristics are only applicable to the variable sized bin packing.

*1) Assignment:* Bin assignment is independent of bin allocation. Well known bin assigment heuristics also known as fitting include best fit, first fit, next fit and worst fit [16].

- Best Fit. The algorithm keeps bins open even if the next item in the list does not fit in previously opened bins, with the expectation that a later smaller item will fit. The criterion for assignment is the placement of the next item.
- First Fit. The algorithm places the next item in the list into the first bin which has space to accommodate the item. When bins are filled completely they are closed and if an item can not fit into any currently open bin, a new bin is opened.
- Next Fit. The algorithm opens a bin and places the items into it in the order they appear in the list. If an item on the list does not fit into the open bin, we close this bin permanently, open a new one, and continue packing the remaining items on the list.
- Worst Fit. The algorithm places the item into a currently opened bin with the most room remaining. If there is no such a bin, a new bin is opened for the item.

*2) Allocation:* The bin allocation can be lazy or aggressive, depending upon where the allocation occurs in the context of the algorithm. The heuristics we introduce here are aggressive and allocations are determined during packing run time execution. This differs from the lazy allocation strategy in [5], in which all allocations are defaulted to the largest bin and re-allocations happen after the single sized bin packing finishes.

- Best Fit. Given the current item, the algorithm opens the bin that best fits the item with least space wasted. The probability of packing smaller items together later is low.
- Expect Fit. The algorithm looks at all the items that have not been packed and allocates a bin that can fit all of them with the least space wasted. If the size of all these items is larger than the largest bin, it allocates the largest bin to the current item. It is highly possible that smaller items can be packed into larger bins that were allocated earlier.

Table II
HEURISTICS SPACE

| Assignment \ Allocation | Best Fit | Expect Fit |
|---|---|---|
| Best | 0 | 1 |
| First | 2 | 3 |
| Next | 4 | 5 |
| Worst | 6 | 7 |

We list the heuristics in Table II and label them by an ID number from 0 to 7. There are more heuristics, such as sum of subsets and a generic algorithm [17] that can be added into the heuristics set. In this paper, the size of the heuristic set is enough to demonstrate how we explore the relationship between optimization decisions and the bin packing features to achieve high quality bin packing dynamically.

In this work, we build a prediction model to map bin packing features to the heuristic which yields a better quality solution defined by the cost function. The optimization decision is made by the observation of the features of bin packing from both bins and items and the learned relation model. We list the procedure in Algorithm 1.

---

**Algorithm 1** Optimization Decision

1: $< i, b > \leftarrow features(Items, Bins)$
2: $s \leftarrow predict(M, < i, b >)$
3: $< assignment, allocation > \leftarrow Heuristic\ Space(s)$
4: **if** $< i, b > \notin DataSet$ **then**
5: $\quad \vec{P} \leftarrow offline\ Profiling(< i, b >)$
6: $\quad s \leftarrow Labeling(\vec{P})$
7: $\quad M \leftarrow < s, < i, b >>$
8: **end if**
9: $RBins = \emptyset$
10: **while** $Items \neq \emptyset$ **do**
11: $\quad item \in Items$
12: $\quad Items = Items - \{item\}$
13: $\quad bin = Assign(item, RBins, assignment)$
14: $\quad$ **if** $bin \neq \emptyset$ **then**
15: $\quad\quad bin \leftarrow item$
16: $\quad$ **else**
17: $\quad\quad bin = Allocate(item, Bins, allocation)$
18: $\quad\quad RBins = RBins \cup \{bin\}$
19: $\quad$ **end if**
20: **end while**
21: **return** $RBins$

---

The algorithm first extract the features $< i, b >$ from the bins and items for the packing instance. Then it applies the learned prediction model $M$ to predict the best packing strategy $s$. The strategy is mapped to a pair of heuristics, $< assignment, allocation >$ defined in Table II. The packing algorithm will pack the item using the selected packing strategy. It assigns bins and allocates a proper bin from the available bin set $Bins$ when necessary. $DataSet$ contains the item/box dimension data and the performance data for different packing heuristics.The algorithm will measure whether the new instance is close to those inside the $DataSet$. If not, it triggers a profiling run and adds its feature and label into the data set. This way we can keep the $DataSet$ up to date.

## IV. PREDICTION MODEL

The purpose of the prediction model is to catch the internal connections between the bin packing features and the selection of heuristics. This task can be achieved by traditional AI classification algorithms such as decision trees, neural nets, bayesian classifiers, support vector machines, and cased-based learning. There are advantages of using these algorithms. The models are simple and in most cases they are easy to interpret and explain.

However, these shallow structures heavily rely on the quality of data engineering to extract features. Better data often beats better algorithms. On one hand, the learning complexity grows exponentially with linear increase in the dimensionality of the data which leads to the curse of dimensionality [10]. On the other hand, if incomplete or erroneous features are extracted, the classification process is inherently limited in performance [11]. Deep learning structure can mitigate these issues by letting the raw input data with noise propagated and filtered in the multi-level deep structure.

### A. Deep Learning

In big data era, deep learning [11] [18] is widely applied to solve artificial intelligence machine learning problems. It demonstrates impressive performance to handle large scale data in computer vision, speech recognition, natural language processing, finance and advertisement etc.

Deep learning is defined by a multiple layered artificial neural network structure and a set of algorithms to train the parameters of the network. The network essentially is cascades of parameterized non-linear modules at each network layer. The training process is to find the parameters of non-linear modules that can minimize a specific loss function computed on the training data set.

A non-linear module is defined by a non-linear function called activation function. The non-linear modules give the network the ability to fit a complex non-linear relation. A typical non-linear function is $sigmoid$ which map numbers in the range of $(-\infty, +\infty)$ to numbers in the range of $(0, 1)$. The loss function could be as simple as the mean square error function for regression. For classification, categorical cross entropy function has better performance. Backward propagation of errors from the loss function, called backpropagation [11] [18] algorithm, is commonly used to train artificial neural networks. An optimization method such as gradient descent is used to search for the parameter values. We will formally define the network structure for our experiment later.

### B. Data Collection

To evaluate our proposed approach, we collect a 4 million data set by randomly sampling real world customer produced logistics orders of Walmart e-Commerce. It has totally $4,278,645$ instances of the item and box dimension data. Each instance contains a set of items and a set of boxes to construct a pack. We keep only instances containing more than one items per order and having with multiple box options to construct the learning model. For each instance, we have an off-line profiling run across all the heuristics which are listed in Table II. We label each instance with the heuristic label using our labeling algorithm.

### C. Features Extraction

We incorporate prior knowledge to speed up the learning procedure and increase the prediction accuracy by the predefined features. Instead of applying dimension reduction processing like PCA [6], Factor Analysis [7] and NMF [19], we use all the raw features listed in Table III. The first seven features are from the items and the next six features are from the bins. The last three features describe the interaction between the items and bins.

Table III
FEATURES

| feature type | features |
|---|---|
| Item features | number of items<br>sum of the item volumes<br>min item volume<br>max item volume<br>average item volume<br>standard deviation of item volumes<br>variance of item volumes |
| Box features | number of boxes<br>min box size<br>max box size<br>average box size<br>standard deviation of box sizes<br>variance of box sizes |
| Cross features | average fill in ratio<br>average min fill in ratio<br>average max fill in ratio |

### D. Performance Signature and Instance Labeling

The raw performance for each packing instance is a vector of packing cost results, $RP$. Each element of the vector represents the packing cost using a different heuristic. We first normalize each raw performance vector to $NP$ (Normalize Performance vector) using equation 1. From the normalized performance vector, we derive the performance signature, $SP$ using equations 2, 3 and 4. The performance signature vector is a binary indicator vector. The non zero elements indicate the top performers within the candidate heuristic set. The $\Delta$ give some tolerance to extract top performers. Smaller value will lead to a larger number of different signatures.

$$NP = \frac{RP}{\sqrt{\sum (RP_i)^2}} \quad (1)$$

$$NP_{min} = min(NP_0, NP_1, ..., NP_7) \quad (2)$$

$$SP = [sp_0, sp_1, ...sp_7] \quad (3)$$

$$sp_i = \begin{cases} 1, & \text{if } NP_i \leq NP_{min} + \Delta \\ 0, & \text{otherwise} \end{cases} \quad (4)$$

The performance signature, $SP$ is generated by locating the lowest cost heuristics in the normalized cost vector with a tolerance factor $\Delta$. The labeling is to map a signature vector which might contain multiple ones or top performers to an indicator vector which contains only one top performer. The reason why we do not measure the raw vector distance is because we care about the right decision to choose the best performers instead of knowing their actual distance.

The challenge of labeling a large set of binary vector lies in two folds: the large overlap of vector indicators between vectors and the computational complexity to label them.

In most cases a performance signature vector is not an indicator vector with a single indicator. The true top performer for a specific instance is either buried in the noises or there are multiple heuristics generating the same result. It will lead to low accuracy when we build the prediction model, if we mislabel the instance. To overcome this challenge, we need to first identify the clusters from the data set. Then, given a cluster, identify the indicator that this cluster represents.

There are many ways to cluster the binary vector. The essential of the clustering is the definition of the distance measurement [20]. One recent approach is entropy based clustering [13], which actually measures the KullbackLeibler (KL) distance and minimize the average KL distances among clusters. The drawback of this approach is its time complexity. It is Monte Carlo based random algorithm to split the original data set into multiple clusters. Given a large binary set, it takes too many iterations or a long time to generate the clusters. A proper initialization of clusters can reduce the complexity of this problem.

Instead of using entropy based clustering directly to cluster the original large binary vector set, we in this work first apply a fast approach to cluster the original data by differentiating leading zeros prefixes. According to the number of leading zero prefix from 0 to 7, we have 8 difference labels. We show the procedure in figure 1. Each dark node is a label class representing a different top heuristic. The index of the first non-zero element represents the top heuristic in

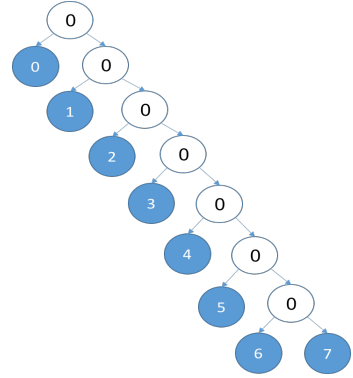

Figure 1. The Prefix Based Clustering

Table II. This way we will get both the cluster and labels. The entropy of the generated clusters are not balanced and clusters with small ID number have larger entropy values. We then apply the entropy based Monte Carlo to refine the labeling results.

## V. SYSTEM OVERVIEW

We show the high level overview of the system architecture in figure 2 and we also briefly introduce experiment environment in this section. The overall flow is as follows. Given a packing instance, it will be sampled before we obtain its profiling performance across different heuristics and add its profiling execution results into the data set. Depending on the instance characteristic, it can query the model for optimized heuristic or directly go to the packing algorithm component if it is a very simple instance such as one item instance. Thus, the system achieves dynamic bin packing.

For an incoming instance, the sample and profiling component makes a decision whether to add this instance to the data set and have a profiling run to obtain the best cost heuristic label. There are reasons why this step is critical. First, it filers out simple cases such as one item packing and keep the instances in data set balanced and less noisy. Second, the profiling run incurs overhead on computation resource. This component can prevent unnecessary computation from happening.

The prediction model is the output of the deep neural network training. The training is computationally intensive and usually it takes about an hour to run on a four million record data set in each epoch. We will explain more details about neural network structures in the next section.

The packing algorithm component embeds strategies to handle different heuristics. Besides the heuristics listed in Table II, we have a set of rule heuristics to deal with simple packing cases like one item packing to speedup the overall system performance.

The system is designed to run on commodity hardware. We conduct our experiment on a virtual machine with 32

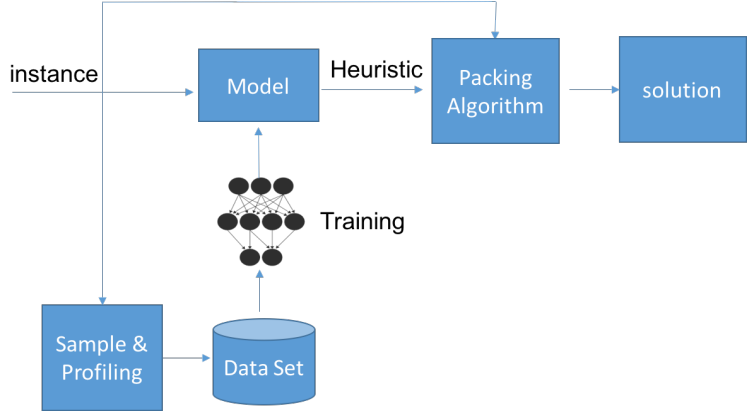

Figure 2. The System Architecture Overview

cores running CentOS 6.7 Linux. It has 32 Intel Xeon E312xx cores running at 2394 MHz. Each network training instance requires six cores running in parallel and this VM can support us to search the tunning space by running five to six training instances at the same time. We build the packing algorithm component using Oracle Java 8 and the deep learning model using tensorflow 0.7.1 [21] with Keras version 1.08 [22] as the front end.

## VI. NEURAL NETWORK

The prediction model in our paper is a supervised machine learning model which is built on top of a deep neural network structure. Formally, given a set of responses $Y$ and observations $X$, the supervised learning task is to estimate a predictor $\hat{Y} = F(X)$. The prediction, $\hat{Y}$ via this predictor is close to the original $Y$. This distance is measured and optimized by a loss function $L(\hat{Y}, Y)$.

In the context of a deep learning task, given a neural network with L layers, the layer $l \in \{0, 1, 2, ..., L-1\}$ is modeled by a set of weight parameters $W_l$ which is a $m \times n$ matrix and bias parameters $b_l$ which is a $n$ dimension vector, and an activation function $f_a$. The two parameters linearly transforms $n$ input values to $m$ outputs and the activation function will map these $m$ outputs into a scalar real value. The $m$ number of outputs are decided by the number of nodes in the next level $l+1$. This value can be different at each level. $f_a \in \{softplus, relu, tanh, sigmoid, linear...\}$ can be any implementation. In our work, we use the same type of $f_a$ at each layer. The output of the layer $l$ is defined in equation (5). The whole network is defined in the composition of a set of equations from (5) to (7). When $l = 0$, we have a special input layer defined in (7).

$$Y_i^{(l)} = f_a(W_l \cdot Y^{(l-1)} + b_l) \quad (5)$$
$$Y^{(l)} = [Y_1^{(l)}, Y_2^{(l)}, Y_3^{(l)}, ..., Y_m^{(l)}] \quad (6)$$
$$Y^{(0)} = X = [x_1, x_2, x_3, ..., x_m] \quad (7)$$

### A. Network Structure

Our network instance is defined by (8) to (12) based on equation (5) to (7). $m_i$ represents the number of nodes at level $l$.

$$L = 6 \quad (8)$$
$$M = [128, 128, 128, 128, 128, 1] \quad (9)$$
$$f_{a(l)}(x) = softplus(x) = \log_e(1 + exp(x)), l \leq 5 \quad (10)$$
$$f_{a(l)}(x) = softmax(x) = \frac{e^{z_j}}{\sum_{k=1}^{K} e^{z_k}}, l = 6 \quad (11)$$
$$L(\hat{Y}, Y) = \sum_{k=1}^{K} (\hat{y}_k log(y_k) + (1 - \hat{y}_k) log(1 - y_k)) \quad (12)$$

We apply softplus as the activation function to all the hidden nodes except for the last layer where we apply softmax, as it is designed to model a classification predictor. Equation (12) defines the loss function. It is the cross-entropy between the network's predicted responses and the real labels. We use Adammax as the optimizer to train the network.

### B. Speed Up Learning by Normalization

During the training of a multiple layer deep neural network, a problem called internal covariate shift may happen [23]. It changes the distribution of the network activations. It is caused by the changing of network parameters during the training process. This problem will slow down the training process by requiring lower learning rates and more careful parameter selection. The work in [23] proposed a normalization for each layer outputs during each training mini-batch. It can speedup the steps to converge. We apply the normalization at each layer to speedup training.

### C. Over Fitting

The neural network has a large number of parameters to estimate during the training. It is very easy to overfit the training data set such that the performance on the testing data set is not as good as that on the training data set. One way to prevent the overfitting is to combine the predictors of many different neural network models. The random drop [24] is a technique to construct many different sub network models by randomly dropping out some network nodes from the the initial network. The final predictor is the average of these sub network predictors.

$$d^l = Bernoulli(p) \quad (13)$$
$$Y^{(l-1)}_{dropOut} = Y^{(l-1)} \cdot d^l \quad (14)$$
$$Y^{(l)}_i = f_a(W_l \cdot Y^{(l-1)}_{dropOut} + b_l) \quad (15)$$

Formally, at each network layer, we introduce a random vector $d^l$ defined in (13). Each of its element is an independent Bernoulli random variable with the possibility of $p$ to be 1. The drop out vector will be applied to outputs for each of the network layer by performing a element wise product defined in (14). Now the composition network structure definition (5) becomes (15) with the drop out implementation. In other words, $d^l$ is used to sample many sub-networks of a large network. The training and updating process is to average all these sub-networks.

## VII. EXPERIMENT RESULT

We in this section first analyze the characteristics of data by showing how we label the data. We also exam the effectiveness of our labeling algorithm. Then, we report the results of our prediction model for classification. The objective of the evaluation results in this section is to demonstrate how our proposed prototype system can improve 1D variable sized bin packing via deep learning.

For each of the four million instances, we have profiling runs for all of the eight heuristics and obtain the performance signatures for each of the instance. We apply the prefix based clustering to create eight clusters. Each represents a different label. We show the signature distribution for each cluster in figure 3. The y-axis represents the different clusters or labels. The x-axis represents the set of all binary signatures with 8 bits. The binary signatures are ordered in the grey code order, so that the adjacent binaries are very similar and only have one bit difference. The circle areas represent the distributions of binary signatures by the normalized counts. The bigger the area, the larger the probability. For all of the eight clusters, one or two signatures will dominate each cluster. That is what our experiment prefers.

In figure 4, we show how the Monte Carlo can help to reduce entropies of the clusters generated by the prefix based fast clustering. The y axis represents the weighted average binary entropy across the eight clusters. The x-axis represents the number of iterations. The entropy decreases linearly with the number of iterations. However, we find the progress is very slow when we try to handle a large set of binaries.

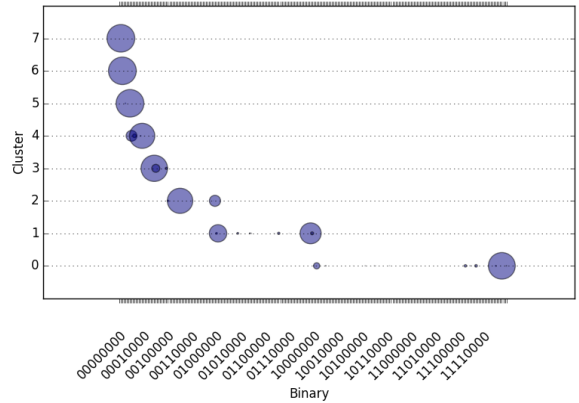

Figure 3. Binary Distribution for Each Cluster/Label

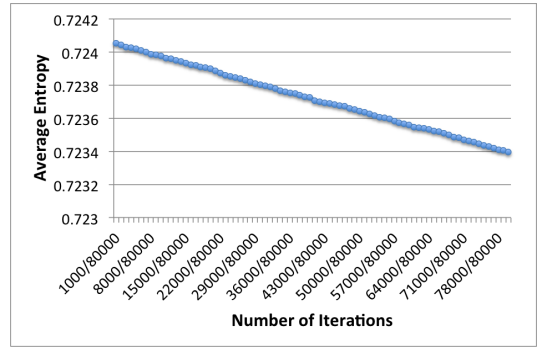

Figure 4. Clustering Entropy During Monte Carlo

In figure 5, we show the break down of the eight heuristics using fast prefix based labeling. The position of the first non zero indicator in the performance signature determines the heuristic label for the instance. This break down gives us a initial overview of the data. $98\%$ instances are dominated by the first four heuristics. The rest four heuristics outperform in only around $2\%$ of all the instances. The instance data set is significantly imbalanced. On one hand, a classifier trained from an imbalanced data set is highly possible to be biased towards the classes with more instances and show poor classification accuracy on the minority classes. On the other hand, this imbalance means the chance to encounter these instances is small and the penalty of not catching the best heuristics can be ignored. Thus, we ignore the rest four heuristics from our perdition model.

In figure 6, we show our neural network model's training accuracy and loss during the training epochs for the four dominated labels: 0, 1, 2 and 3. This figure shows how our model is built up on the training data set. After the first several epochs, the converge speed becomes slow. This is because we have a large number of nodes at each level to train.

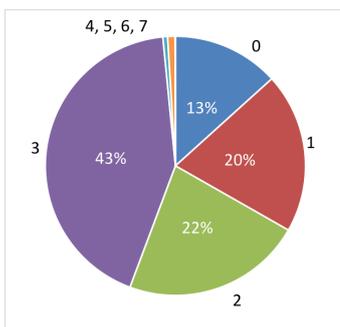

Figure 5. Instance Distribution by the Top Performance Heuristics

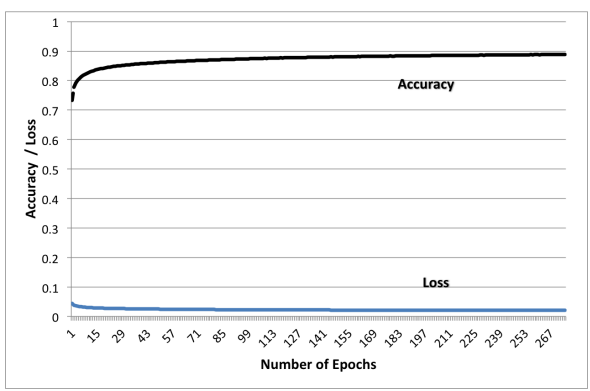

Figure 6. The Loss and Accuracy during Training Iterations

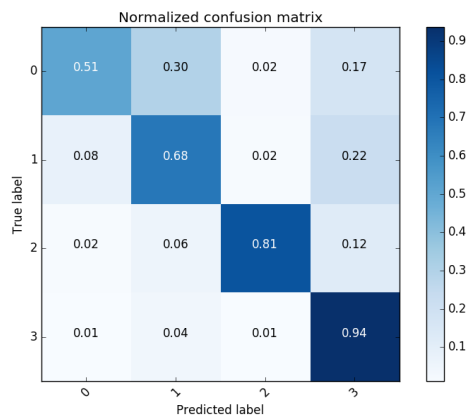

Figure 7. The Prediction Results in a Confusion Matrix

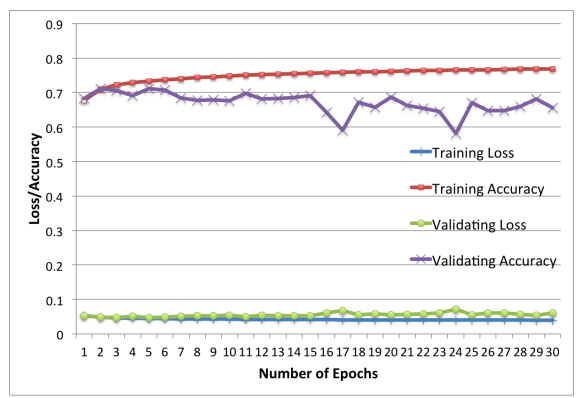

Figure 8. Train vs. Validation Loss and Accuracy

We show our model prediction on the training data set by a normalized confusion matrix in figure 7. Our classification model is trained to classify the four dominated labels, the normalized confusion matrix summarizes the ratios of false positives, false negatives, true positives, and true negatives. The x axis represents the predicted labels and the y axis represents the true labels. Each row of the matrix represents the distribution of the predicted values. Each element at row i and column j means for the true label i what is the percentage that the predicted label j is classified as the label i. The matrix diagonal are all true positives. We can see that the smaller labels have smaller true positives. This is because the way we create labels. The smaller label have larger binary entropy in its cluster. By running longer time, our Monte Carlo process decreases the overall entropy for all the clusters, and thus our model can improve these smaller true positives.

In the figure 8, we show our model's capability of prediction by comparing the training accuracy/loss and validating accuracy/loss. The original data set are split into a 5% validating data set and an 95% of the training data set because we have a large date set. We only show the results during the first 30 epochs because the converge will become slow after the first 30 epochs. We can see that over-fitting happens soon. Most validating accuracy is mostly between 60% to 70%.

## VIII. RELATED WORK

E. Lopez et al. in [6] focus on the feature engineering for fixed sized 1D and 2D bin packing problems. Their paper uses a dimension reduction technique, Principal Component Analysis (PCA) as a knowledge discovery method. Their work tries to gain a deeper understanding of the structure of bin packing problems and how this relates to the performance of heuristic approaches applied to solve them. The PCA technique gives a valuable indication of the combination of features characterizing with how easy or hard to solve instances. They found that there exists correlations between instance features and heuristic performance.

J. Perez O. et al. [7] propose a methodology to model the algorithm performance predictor. Given a set of solved instances of a NP-hard problem and set of heuristic algorithms, the authors model the relationship among performance and characteristics to predict which algorithm solves the problems better. The relation is learned from historical data using machine learning techniques. Specifically, it is a unsupervised clustering technique. They benchmark their model using the classical bin packing problem (fixed sized bin packing). They apply factor analysis to engineer features that represent critical characteristics of the fixed size bin packing problem. They develop predictors that incorporate the interrelation among five critical characteristics and the performance of seven heuristic algorithms. They claimed that they obtained an accuracy of 81% in the selection of the best algorithm.

K. Sim, et al. in [8] are inspired by the immune systems and propose a novel hyper-heuristic system that continuously learns over time to solve a combinatorial optimization problem. They tested their system on a large corpus of 3,968 new instances of fixed size 1D bin-packing problems. Compared with the previous two works [6] [7] that have a fixed set of heuristics, their system continuously generates new heuristics and samples problems from its environment. Thus representative problems and heuristics are incorporated into a self-sustaining network of interacting entities inspired by methods in artificial immune systems. It is capable of generalizing over the problem space.

Compared with the above related works [6] [7] [8] which predict the end to end performance using different heuristics to perform the end to end computation, P. Ross etc. in the work [9] optimize different phases of the execution for the bin packing problem computation. The authors open the evolutionary algorithms(EAs) as an white box. In contrast, the existing approaches neither offer worst-case bounds, nor have any guarantee of optimality when used to solve individual problems. They can also take much longer than non-evolutionary methods. The authors use a learning classifier system XCS to learn a solution process rather than to solve individual problems. The process chooses one of various simple non-evolutionary heuristics to apply to each state of a problem and gradually transforms the problem from its initial state to a solved state. The authors test their solution on a large set of fixed size one-dimensional bin packing problems.

The above related works apply traditional machine learning to solve the fixed one dimensional bin packing problem. The quality of the prediction model largely depends on the feature engineering [11] [18]. Deep learning can mitigate this feature engineering issue. Majority applications of deep learning are in the domain of image processing. The follow related works apply it in non image processing domains.

Zhang et al. in [25] predict user responses, such as click-through rate and conversion rate for web applications such as web search, personalized recommendation, and online advertising. Unlike continuous raw features in the image and audio domains, the input features in web space are always of multi-field and are mostly discrete and categorical while their dependencies are little known. The traditional model builds up high-order combination features. To tackle the issue, the authors propose two novel models using deep neural networks to automatically learn effective patterns from categorical feature interactions and make predictions of users ad clicks. They claim that large-scale experiments with real-world data demonstrate their methods work better than major state-of-the-art models.

The work in [19] propose a robust classifier to predict buying intentions based on user behavior within a large e-commerce website. The authors compare traditional machine learning techniques with the most advanced deep learning approaches. The results show that both Deep Belief Networks and Stacked De-noising auto-Encoders achieved a substantial improvement by extracting features from high dimensional data during the pre-train phase. They also prove that it is more convenient to deal with severe class imbalance.

In ths paper [26], J. Saxe et al. introduce a deep neural network based malware detection system. They claim that it achieves a usable detection rate at an extremely low false positive rate and scales to real world training example volumes on commodity hardware. In addition, they describe a non-parametric method for adjusting the classifiers scores to better represent expected precision in the deployment environment.

N. Polson et al. in [27] develop a deep learning predictor to model and predict traffic flows. The challenge in the work is that traffic flows have sharp nonlinearities resulting from transitions from free flow to breakdown and then to congested flow. Their paper uses a deep learning architecture to capture nonlinear spatial-temporal flow effects. Their paper focus on forecasting traffic flows during special events, such as a Chicago Bears football game and an extreme snowstorm, where the sharp traffic flow regime can occur very suddenly and hardly predictable from historical patterns.

IX. CONCLUSION

In this paper, we focus on optimization of the the 1D variable sized bin packing problem through a deep learning approach. We first define the optimization heuristics space for this particular bin packing problem. Then, we model a large neural network to predict the optimal strategy for each bin packing instance. Our model can automatically capture the features from a large number of raw data sets; our fast label procedure can help to balance the labeling quality and speed. Our predictive system achieves up to 89% training accuracy and 72% validation accuracy. To our knowledge, it is the first exploration of applying deep learning for solving the variable sized bin packing problem.

ACKNOWLEDGMENT

We would like to thank Vadim Zilberleyb for his assistance in providing a sample data set for our initial experiments, Dr. Tony Qin, Dr. Lei Zhang and Dr. Yi Feng for their suggestions on traditional machine learning models, Dr.

Hengheng Chen for her assistance in obtaining a large data set. We are also grateful for the support from Multi Channel Sourcing Engine team and Customer Promise division at Walmart Labs.